\documentclass[sigconf]{acmart}

\usepackage{amsmath}
\usepackage{amsfonts}
\usepackage{amsthm}
\usepackage{algorithm}
\usepackage{algorithmic}
\usepackage{multirow}
\AtBeginDocument{%
  \providecommand\BibTeX{{%
    \normalfont B\kern-0.5em{\scshape i\kern-0.25em b}\kern-0.8em\TeX}}}

\acmConference[XXXXXX 'XXX]{XXXXXX 'XXX: ACM Symposium on XXXX XXXX XXXX }{XXX XX--XX, 20XX}{XXXXXX, XX}
\acmBooktitle{XXXXXX 'XX: ACM Symposium on XXXX XXXX XXXX,
  XXX XX--XX, 20XX, XXXXXX, XX}
\acmPrice{15.00}
\acmISBN{XXX-X-XXXX-XXXX-X/XX/XX}

\begin{document}

% the following package is optional:
%\usepackage{latexsym}

% See https://www.overleaf.com/learn/latex/theorems_and_proofs
% for a nice explanation of how to define new theorems, but keep
% in mind that the amsthm package is already included in this
% template and that you must *not* alter the styling.
%\newtheorem{example}{Example}
%\newtheorem{theorem}{Theorem}

% Following comment is from ijcai97-submit.tex:
% The preparation of these files was supported by Schlumberger Palo Alto
% Research, AT\&T Bell Laboratories, and Morgan Kaufmann Publishers.
% Shirley Jowell, of Morgan Kaufmann Publishers, and Peter F.
% Patel-Schneider, of AT\&T Bell Laboratories collaborated on their
% preparation.

% These instructions can be modified and used in other conferences as long
% as credit to the authors and supporting agencies is retained, this notice
% is not changed, and further modification or reuse is not restricted.
% Neither Shirley Jowell nor Peter F. Patel-Schneider can be listed as
% contacts for providing assistance without their prior permission.

% To use for other conferences, change references to files and the
% conference appropriate and use other authors, contacts, publishers, and
% organizations.
% Also change the deadline and address for returning papers and the length and
% page charge instructions.
% Put where the files are available in the appropriate places.

\title{Black-box Gradient Attack on Graph Neural Networks: Deeper Insights in Graph-based Attack and Defense}

\author{Haoxi Zhan}
\email{zhanhaoxi@foxmail.com}
\author{Xiaobing Pei}
\authornote{Corresponding Author}
\email{xiaobingp@hust.edu.cn}
\affiliation{%
  \institution{Huazhong University of Science and Technology}
  \city{Wuhan}
  \state{Hubei}
  \country{China}
}

% Multi author syntax
%\author{
%Haoxi Zhan$^1$\and
%Xiaobing Pei$^1$ \footnote{Corresponding Author}\\
%\affiliations
%$^1$Huazhong University of Science and Technology\\
%\emails
%zhanhaoxi@foxmail.com,
%xiaobingp@hust.edu.cn
%}

\begin{abstract}
Graph Neural Networks (GNNs) have received significant attention due to their state-of-the-art performance on various graph representation learning tasks. However, recent studies reveal that GNNs are vulnerable to adversarial attacks, i.e. an attacker is able to fool the GNNs by perturbing the graph structure or node features deliberately. While being able to successfully decrease the performance of GNNs, most existing attacking algorithms require access to either the model parameters or the training data, which is not practical in the real world.

In this paper, we develop deeper insights into the Mettack algorithm, which is a representative grey-box attacking method, and then we propose a gradient-based black-box attacking algorithm. Firstly, we show that the Mettack algorithm will perturb the edges unevenly, thus the attack will be highly dependent on a specific training set. As a result, a simple yet useful strategy to defense against Mettack is to train the GNN with the validation set. Secondly, to overcome the drawbacks, we propose the Black-Box Gradient Attack (BBGA) algorithm. Extensive experiments demonstrate that out proposed method is able to achieve stable attack performance without accessing the training sets of the GNNs. Further results shows that our proposed method is also applicable when attacking against various defense methods.
\end{abstract}




\maketitle
\section{Introduction}

Graph structured data is widely used in a variety of domains, such as social networks\cite{social}, academic publishing\cite{semantic}, recommender systems\cite{recom} and financial transactions\cite{trans}. How to learn effective graph representations has long been an important research direction. Recently, Graph Neural Networks (GNNs) has become the mainstream method for graph representation learning\cite{sunsurvey}. Firstly introduced by \cite{firstmodel}, the Graph Convolutional Network (GCN)\cite{kipf} achieved state-of-the-art performance in the node-classification task and is considered as the most representative GNN model. Then, numerous GNN models, such as GAT\cite{GAT}, GraphSAGE\cite{GraphSAGE} and JK-Net\cite{JK}, have been proposed.

It has been revealed that deep learning models are often in lack of robustness\cite{lack1}. It is possible to fool the model by generating perturbations deliberately. GNNs are no exceptions. Adversarial attacks on GNNs could be classified using  various criteria. According to the attackers' knowledge, we could divide existing GNN attacks into white-box attacks, grey-box attacks and black-box attacks\cite{zssurvey}. In the white-box scenario, the adversary has access to both the model parameters and the training data, which even include the ground-truth labels. Grey-box attacks only need partial information and they are more practical than the white-box ones. Representative grey-box attacking methods include Nettack\cite{nettack}, which aims to misclassify a set of targeted nodes, and Mettack\cite{mettack}, which aims to reduce the overall classification accuracy rates. Since model parameters are not accessible in grey-box scenarios, a surrogate model is usually trained on the training set to approximate the gradients of potential perturbations. Black-box attacks allow the most limited knowledge such that only black-box queries are possible\cite{chisurvey}. Since the access to the training set is also prohibited, reinforcement learning is introduced to perform black-box attacks\cite{msusurvey}. However, the queries are still considered to be impractical and RWCS, a black-box attack based on the theory of random walk, has been proposed\cite{rwcs}.

Various defense methods have also been proposed. For instance, GCN-Jaccard \cite{jaccard} increases the robustness of GCNs by eliminating edges with low similarity before the training process. R-GCN \cite{rgcn} utilizes the attention mechanism which regards node features as Gaussian distributions and assigns attention scores according to the variances. By introducing structure learning, Pro-GNN\cite{prognn} achieves promising results when defending against structure perturbations.

Although defense algorithms have emerged to enhance the security of GNNs, studies on deep learning models show that it is possible to improve the adversarial attacks to degrade the defense performance\cite{obfus,noteasy}. Meanwhile, in-depth studies on adversarial attacks is an efficient tool to develop insights in deep learning models\cite{lack2}. Comparing with Deep Neural Networks (DNNs), adversarial attacks on GNNs remain poorly understood.

One perspective to improve adversarial attacks is to understand how existing attacks work at first. In this paper, we target on investigate Mettack, which is the most representative non-targeted grey-box attack, to develop a deeper understanding of graph attack and defense. Our case studies show that the perturbations chosen by Mettack are highly correlated with the training set and they are distributed unevenly on the graph. As a result, we reveal that a simple defense strategy is to train the GNN with the validation set. Based on the findings, we propose the Black-Box Gradient Attack (BBGA) algorithm to evenly perturb the graphs without accessing any ground-truth labels. Extensive experiments on three real-world benchmark datasets demonstrate that our proposed method can effectively attack the graph structure without accessing any training labels.

In summary, our contributions of this paper are as follows:
\begin{itemize}
\item {\bf How Mettack works}: We studies the patterns of Mettack perturbations and we show that the perturbations are denser near the training set. We show that such unevenness helps in grey-box attacks but it could also be utilized by defenders.
\item {\bf Black-box Gradient Attack (BBGA)}: While it is believed that training surrogate model without ground-truth labels are impossible, we propose the first gradient-based black-box attacking method to our best knowledge. It is also the first non-targeted graph structure attack without permission to do black-box queries. Exploiting the spectral clustering, we train the surrogate model with pseudo-labels and then evenly distribute the perturbations via a novel $k$-fold training strategy. 
\item {\bf Extensive experiments}: We conduct numerous experiments to show the effectiveness of our proposed method. Further studies are provided for the explanability of the method.

\end{itemize}

The rest of our paper is organized as follows. In Section 2, we define the mathematical notations used in this paper and then we introduce the preliminaries. In Section 3, we analyse Mettack perturbations via a series of case studies. In Section 4, we introduce our propose BBGA method and explain it in details. Experimental results, ablation studies and parameter analysis are reported in Section 5. Finally, we conclude the paper in Section 6.

\section{Preliminaries}

In this section, we introduce the notations used in this paper as well as fundamental concepts.

\subsection{Mathematical Notations}

In this paper, we denote a graph with $N$ nodes as $G=(V,E)$ where $V = \{v_1, \cdots, v_N\}$ is the node set and $E = \{e_1, \cdots, e_E\} \subseteq V\times V$ is the edge set. The set of vertices $V$ is usually divided into the training set $V_{\text{train}}$, the validation set $V_{\text{val}}$ and the testing set $V_{\text{test}}$. We denote the structure of the $G$ via the adjacency matrix $A \in \{0,1\}^{N\times N}$ whose element $A_{ij} = 1$ if and only if $\exists i\in [1,E] \text{ s.t. } e_i\in E$ connects $v_i$ and $v_j$. Node features are stored in a matrix $X \in \mathbb{R}^{N\times F}$ where $F$ is the number of dimensions of node features. The labels of a dataset is denoted by $C$ while pseudo-labels used during training are denoted as $C_p$.

\subsection{Graph Convolutional Networks(GCNs)}
Despite a number of GCN models have been proposed, in this paper we mainly consider the most representative one introduced by \cite{kipf}. Each layer of a GCN aggregate messages according to the graph structure and then perform linear transformation on the node features. Such a graph convolutional layer could be denoted as the following equation:
\begin{equation}
H^{(l+1)} = \sigma (\hat{A} H^{(l)} W^{(l)}),
\end{equation}
where $\hat{A} = \tilde{D}^{-\frac{1}{2}} \tilde{A} \tilde{D}^{-\frac{1}{2}}$ is the normalized adjacency matrix such that $\tilde{A} = A + I_N$, $\tilde{D}_{i,i} = \sum_{j} \tilde{A}_{i,j}$. $\sigma$ is a non-linear activation function. A typical GCN network consists of two layers, the whole network is usually described as:
\begin{equation}
Z = f(X,A) = \text{softmax}(\hat{A} \sigma(\hat{A} X W^{(0)}) W^{(1)}).
\end{equation}

The GCN network is usually trained with a cross entropy loss. Noticing that two-layer GCNs aggregate information within 2-hop neighborhoods for each node, a simplified and linearized version is usually used in adversarial attacks:
\begin{equation}
Z = f(A, X) = \text{softmax}(\hat{A}^2 XW).
\label{surrogate}
\end{equation}

To utilize GCN in inductive settings, \cite{GraphSAGE} introduced a variant of GCN with a different normalization method:
\begin{equation}
h_v^{(l)} = \sigma \left(W_{l} \cdot \frac{1}{\tilde{D}_{(v,v)}} \sum_{u\in \tilde{\mathcal{N}}(v)} h_u^{(l-1)}\right),
\label{sage}
\end{equation}
where $h_v^{(l)}$ is the hidden representation of node $v$ in the $l^{th}$ layer and $\tilde{\mathcal{N}}(v)$ is the neighborhood of node $v$ in the self-looped graph with $\tilde{A}$ as the adjacency matrix.

\subsection{Mettack}

Mettack\cite{mettack} is a gradient-based grey-box graph attacking algorithm. Being directly connected to the training loss, gradients are widely used in deep learning adversarial attacks\cite{adv}. Being denied to access model parameters, Mettack trains the surrogate model described in Eq. \ref{surrogate} to approximate the gradients of the GCN model. In order to find the best edge to perturb, the adjacency matrix is regarded as a hyperparameter and the meta-gradients are computed after the training of the surrogate model:
\begin{equation}
\nabla_G^{\text{meta}} := \nabla_G \mathcal{L}_{\text{atk}} (f_{\theta^{*}} (G))
\text{ s.t. } \theta^{*} = \text{opt}_{\theta} (\mathcal{L}_{\text{train}} (f_{\theta}(G))),
\end{equation}
where $\mathcal{L}_{\text{atk}}$ is the target function that the attacker aims to optimize, $\text{opt}$ is a training procedure and $\mathcal{L}_{\text{train}}$ is the training loss. It is revealed that when $\mathcal{L}_{\text{atk}} = -\mathcal{L}_{\text{self}}$, where $\mathcal{L}_{\text{self}}$ is the cross-entropy loss on the unlabelled nodes with predicted pseudo-labels, the algorithm reaches its best performance.

A greedy algorithm, which chooses exactly one edge a step, is employed to perform perturbations. The score of a node pair $(u,v)$ is defined as:
\begin{equation}
S(u,v) = \nabla_{A_{uv}}^{\text{meta}}\cdot (-2 \cdot A_{uv} + 1),
\end{equation}
where $A$ is the adjacency matrix. The signs of meta-gradients are flipped for connected node pairs to yields the gradients for removing the edge.   In each iteration, the algorithm picks the potential perturbation with the highest score.

The total number of perturbations is controlled by a budget constraint $\Delta$, which is usually defined via a perturbation rate $\delta = \frac{\Delta}{|E|}$.

\section{Case studies on Mettack}
Adversarial attacks generate deliberate perturbations on graph data. Hence, to investigate why adversarial attacks work, it is necessary to analyse the perturbations made by attacking algorithms. Previous studies find out that dissimilar nodes tend to be connected by attackers\cite{jaccard}. In this paper, we focus on the distributions of such carefully-crafted perturbations.

\subsection{Unevenness of perturbations}
Consisting of 2 graph convolutional layers, a typical GCN network is only able to aggregate information with the 2-hop neighborhood of each node. Xu et al. proved the following theorem:
\begin{theorem}[\cite{JK}]
If all paths in the computation graph of the model are activated with the same probability. Given a L-layer GCN with Eq. \ref{sage} as the normalization method, then $\forall i,j \in V$, $\mathbb{E} [\frac{\partial H_j}{\partial X_i}]$ is equivalent to the probability of reaching node $j$ via a k-step random walk starting at node $i$.
\label{RW}
\end{theorem}

Inspired by Theorem \ref{RW}, it is nature to raise the question that whether grey-box gradient attacks mainly perturb edges that are adjacent to the training set or not.

\begin{figure}
\centering
\includegraphics[width=0.4\textwidth]{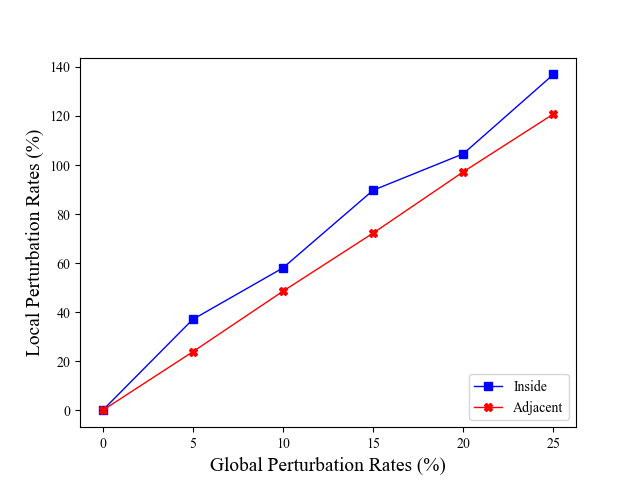}
\Description{The local perturbations rates of mettack perturbed graphs.}
\caption{The local perturbations rates of mettack perturbed graphs.}
\label{fig01a}
\end{figure}

We conducted a case study on the Cora dataset with Mettack. Figure \ref{fig01a} demonstrates the local perturbation rates both inside and adjacent to the training set. As illustrate in the figure, the local perturbation rate inside the training set exceeds $100\%$ when the global perturbation rate reaches $20\%$. Considering the edges $(v_1, v_2)$ that are adjacent to the training set such that
\begin{equation}
(v_1 \in V_{\text{train}} \land v_2 \not\in V_{\text{train}} ) \lor (v_1 \not\in V_{\text{train}} \land v_2 \in V_{\text{train}}),
\end{equation}
the local perturbation rate exceeds $100\%$ when the global perturbation rate reaches $25\%$. Thus, instead of perturbing the edges evenly, Mettack essentially flips a much higher proportion of edges near the training set than the assigned perturbation rate.

\subsection{Ablation Study on The Unevenness}

In order to show the effects of such unevenness, an ablation study on the Cora dataset is conducted. We create two variants of the DICE (Disconnect Internally, Connect Externally) algorithm, which disconnects nodes with the same label and connects nodes with different labels randomly. DICE-Free randomly flips edges on the whole graph. DICE-Control ensures that $10\%$ of the perturbations are within the training set and $90\%$ of the flips are adjacent to the training set. Following \cite{mettack}, we suppose that the attacker knows all the ground-truth labels in this study.

\begin{figure}
\centering
\includegraphics[width=0.4\textwidth]{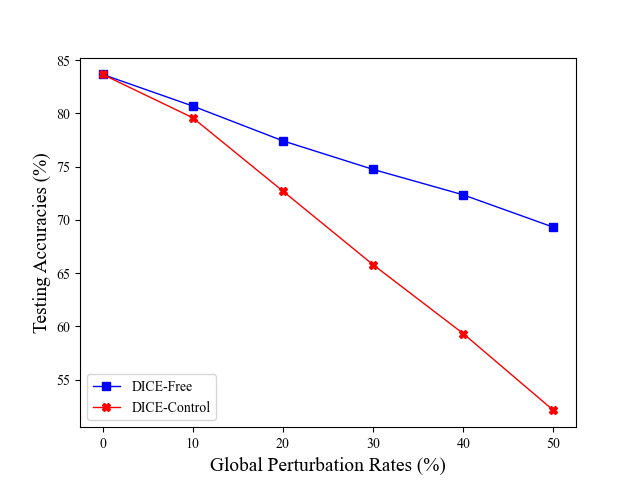}
\Description{The testing accuracies of GCN on DICE-attacked graphs.}
\caption{The testing accuracies of GCN on DICE-attacked graphs.}
\label{fig01b}
\end{figure}

As reported in Figure \ref{fig01b}, the enforced unevenness does not boost the attacking performance too much when perturbation rates are low. However, with the increasing of perturbation rates, DICE-Control outperforms DICE-Free significantly. When the perturbation rates are $50\%$, DICE-Control leads to a $56.05\%$ higher misclassification rate than DICE-Free. The result demonstrates that when the training set is accessible, the unevenness of perturbations helps a lot in grey-box adversarial attacks.

\subsection{Defend Mettack via Validation Set}
\begin{table}
\centering
\begin{tabular}{c | c c }
\hline
Ptb Rate & GCN & Flip-GCN \\ \hline
10\% & 71.59$\pm$1.79 & 81.19$\pm$0.89 \\
20\% & 55.88$\pm$5.03 & 80.50$\pm$0.69 \\
\hline
\end{tabular}
\caption{The experimental results of the flip strategy (Accuracy$\pm$Std).}
\label{flip}
\end{table}
Although the unevenness helps in attacking, it could also be utilized by the defender. We develop Flip-GCN, which is a training strategy that trains the GCN with the validation set, to defend against Mettack. Following the experimental settings of \cite{prognn}, we randomly select 10\% of nodes as the training set, 10\% for validation and the remaining 80\% as the testing set. The graph is perturbed by Mettack without the log-likelihood restraint. As reported in Table \ref{flip}, when $V_{\text{train}}$ and $V_{\text{val}}$ are flipped, the testing accuracy increased dramatically.

\section{Black-Box Gradient Attack}
As demonstrated by the above studies, Mettack has the tendency to perturb the edges unevenly. While such unevenness contributes to the performance, it is in lack of robustness since even training the GCN with the validation set will degrade its performance severely. Meanwhile, the tendency to connect dissimilar nodes are also easily utilized by defense algorithms \cite{prognn,jaccard}.

To increase the robustness of adversarial attacks, the perturbations should be unbiased to any data splits. Thus, indeed the attack is supposed to be black-box to prevent such biases.

While most black-box attacks are based on reinforcement learning \cite{rls2v,rewatt}, random walk-based attacks such as RWCS are also promoted recently to avoid model inquires \cite{rwcs}. However, RWCS focuses on perturbing features instead of graph structures. To our best knowledge, no existing structure-oriented black-box attack works without black-box inquires. Meanwhile, gradients are not exploited in black-box attacks since a surrogate model is needed. However, gradients are powerful tools to pick edges for perturbations. In this paper, by disengaging gradients from any specific training set, we propose the novel Black-Box Gradient Attack (BBGA) algorithm.

\subsection{Attack Condition}
In this subsection, we introduce the attacking condition of our model systematically.

{\bf Goal}: As an untargeted attack, the attacker's goal is to decrease the classification accuracies of GNNs.

{\bf Knowledge}: As a black-box attack, the access to model parameters, training data and ground-truth labels are denied. The graph $G=(V,E)$ and the node features are considered to be accessible.

{\bf Constraints}: The number of changes is restricted by a budget $\Delta$ such that $\lVert A' - A \rVert_0 \leq 2\Delta$. Here $A'$ is the modified adjacency matrix and we have $2\Delta$ due to the symmetry of adjacency matrices. In addition, various defense algorithms have taken advantage of the connections between dissimilar nodes \cite{jaccard,prognn}. We believe that such easily-detected modifications are supposed to be restricted since they could be easily eliminated by a pre-processing algorithm. Noticing that node features in common graph tasks are encoded in the one hot manner, we propose a similarity constraint such that for any pair of nodes $v_1, v_2 \in V$, if their Jaccard similarity score $J_{v_1, v_2}$ is smaller than a threshold value $\eta$, the connection of $v_1$ and $v_2$ is disallowed. The definition of Jaccard similarity score is:
\begin{equation}
J_{v_1, v_2} = \frac{M_{11}}{M_{01}+M_{10}+M_{11}},
\end{equation}
where $M_{ab} (a, b \in \{ 0,1 \})$ represents the number of features which have value $a$ in node $v_1$ and value $b$ in node $v_2$. The constrains are summarized as a function $\phi(G):G\rightarrow \mathcal{P}(V\times V)$, where $G$ is the graph to attack. The function $\phi$ maps the graph to a set of valid node pairs.

\subsection{Pseudo-label and Surrogate Model}
Since no ground-truth labels are accessible, pseudo-labels are needed in order to train the surrogate model. We utilize spectral clustering to generate pseudo-labels\cite{scikit-learn}. Parameters of the  spectral clustering algorithm are chosen according to the Calinski-Harabasz Score.

Since it's not accurate to approximate the gradients of GNNs with pseudo-labels generated by a spectral clustering algorithm, a simplified GCN described in \ref{surrogate} is employed as the surrogate model. The surrogate model is trained with the pseudo-labels on a random training set, which has no relation with the training set of the defense model.

\subsection{$k$-Fold Greedy Attack}
As demonstrated in Section 2, a main drawback of Mettack is the uneven distribution of the modifications. In the black-box scenario in which the training set is not accessible, the attacker is supposed to distribute its modifications in the whole graph. In this paper, we proposed a novel $k$-fold greedy algorithm to solve this problem.

We divide the node set $V$ into $k$ partitions $V_1, V_2, \cdots, V_k$. For the $i^{\text{th}}$ partition $V_i (i\in [1,k])$, we predict $\hat{C}_i$, which is the labels of nodes in $V - V_i$, with a GCN trained on it. The attacker loss function is defined as:
\begin{equation}
\mathcal{L}_{\text{atk}}^{i} = -\mathcal{L}(V - V_i,\hat{C}_i),
\end{equation}
where $\mathcal{L}$ is the cross-entropy loss.

With the attacker's loss, we compute the meta-gradients (gradients w.r.t. hyperparameters):
\begin{equation}
\nabla_G^{i} := \nabla_G \mathcal{L}_{\text{atk}}^{i} (f_{\theta^{*}} (G))
\text{ s.t. } \theta^{*} = \text{opt}_{\theta} (\mathcal{L}_{i} (f_{\theta}(G))),
\end{equation}
where $\mathcal{L}_{i}$ is the cross-entropy loss on partition $V_i$. Similar to Mettack, the partition score function $S^{i}:V\times V \rightarrow \mathbb{R}$ on the $i^{th}$ partition is defined as:
\begin{equation}
S^i (u, v) = \nabla_{u, v}^{i} \cdot (-2\cdot A_{uv} + 1).
\end{equation}

For each pair of node $(u,v)$, we compute $\sigma_{uv}$, which is the standard deviation of its $k$ partition scores. The greedy score for $(u,v)$ is defined as:
\begin{equation}
S(u,v) = \begin{cases}
\sum_{i=1}^{k} S^{i}(u,v), \text{ }& \sigma_{uv} < \tilde{\sigma} \\
0, \quad & (\sigma_{uv} >= \tilde{\sigma}) \lor (u=v) 
\end{cases}
\label{score}
\end{equation}
where $\tilde{\sigma}$ is the median of all $\sigma_{uv}$s. The definition eliminates all self-loops and ensures that the gradients of the chosen perturbation on the $k$ partitions are not too diverse.

In each step, we greedily pick exactly one perturbation $e' = (u',v')$ with the highest score:
\begin{equation}
e' = \mathop{\arg\max}_{e=(u,v)\in \phi(G)} S(u,v),
\end{equation}
where $e\in \phi(G)$ ensures that the modification does not conflict with the attack constraints. Then the algorithm updates $G$ according to $e' = (u', v')$ by flipping the value of $A_{u'v'}$.

\subsection{Algorithm}

\begin{algorithm}[tb]
\caption{Black-Box Gradient Attack (BBGA)}
\label{BBGA}
\textbf{Input}: $G=(V,E)$, node features $X$, attack budget $\Delta$, constraint $\phi(G)$, training iteration $T$, partition $\{V_1, \cdots, V_k\}$. \\
{\bf Output}: Modified graph $G' = (V, E')$.\\
\begin{algorithmic}[1]
\STATE $C_p$ $\leftarrow$ SpectralClustering$(G, X)$.
\STATE $C_s$ $\leftarrow$ training surrogate model with $C_p$ and a random training set.
\STATE $A'$ $\leftarrow$ $A$
\WHILE{$\lVert A' - A\rVert_0 < 2\Delta$}
\FOR{$i \in [1,k]$}
\STATE randomly initialize $\theta_0$;
\STATE $\theta_T$ $\leftarrow$ update parameters for $T$ iterations;
\STATE $\nabla_{G}^{i} \leftarrow \nabla_G \mathcal{L}_{\text{atk}}^{i} (f_{\theta^{T}} (G))$;
\STATE $S_i \leftarrow \nabla_{G}^{i} \odot (-2A' + 1)$;
\ENDFOR
\STATE $S$ $\leftarrow$ according to Eq. \ref{score};
\STATE $e' \leftarrow$ the index of the maximum element in $S$ that are in $\phi(G)$;
\STATE $A' \leftarrow$ flip edge $e'$.
\ENDWHILE
\STATE $G'$ $\leftarrow$ use $A'$ as the new adjacency matrix; \\
{\bf return}: $G'$.
\end{algorithmic}
\end{algorithm}

Following the detailed description, we now present the pseudo-code of the BBGA algorithm in Algorithm \ref{BBGA}.

The meta-gradients of all the $N^2$ pairs of nodes will be computed in the algorithm, the computing of meta-gradients have $T$ steps due to the chain rule and the meta-gradients are calculated for each of the $k$ partitions. Thus, the computational complexity for the attacking procedure itself is bounded by $O(k\cdot T\cdot N^2)$. Considering the computational efforts needed to find the pseudo-labels\cite{fastSC}, the overall computational complexity of BBGA is $O(k\cdot T\cdot N^2) + O(n^3)$.

\section{Experiments}

In this section, we evaluate our proposed BBGA algorithm against both vanilla GCN and several defense methods.We introduce our experimental settings at first and then we aim to answer the following research questions:
\begin{itemize}
\item {\bf RQ1}: How does BBGA works without accessing the training set of the model?
\item {\bf RQ2}: Are the attacked graph as expected such that the perturbations are not distributed mainly around the training set?
\item {\bf RQ3}: How do different components and hyperparameters affect the performance of BBGA?
\end{itemize}

\subsection{Experimental Settings}

\subsubsection{Datasets}
\begin{table}
\centering
\begin{tabular}{|l | c c c c|}
\hline
Dataset & $|V|$ & $|E|$ & Classes & Features \\
\hline
Cora & 2485 & 5069 & 7 & 1433 \\
Citeseer & 2110 & 3668 & 6 & 3703 \\
Cora-ML & 2810 & 7981 & 7 & 2879 \\
\hline
\end{tabular}
\caption{Statistics of datasets. Notice that only the largest connected components of the graphs are considered.}
\label{dataset}
\end{table}

Cora, Citesser and Cora-ML, which are three\linebreak commonly-used real-world benchmark datasets, are employed in our experiments. Following \cite{mettack}, we only consider the largest connected components of the graphs. The statistics of the datasets are summarized in Table \ref{dataset}. Following \cite{prognn}, we randomly pick 10\% of nodes for training, 10\% of nodes for validation and the remaining 80\% of nodes for testing.

\subsubsection{Defense Methods}
To demonstrate the effectiveness of our proposed BBGA method, both vanilla GCN and several defense algorithms are employed in our experiments. For GCN and RGCN we use the official implementation along with its hyperparameters. For other models we use the implementation and hyperparatemers in \cite{deeprobust}.

\begin{itemize}
\item {\bf GCN}\cite{kipf}: In this work, we focus on attacking the representative Graph Convolutional Network (GCN).
\item {\bf GCN-Jaccard}\cite{jaccard}: Noticing that existing graph adversarial attacks tend to connect dissimilar nodes, GCN-Jaccard remove the edges that connect nodes with Jaccard similarity scores lower than a threshold $\eta$ before training the GCN. 
\item {\bf R-GCN}\cite{rgcn}: R-GCN utilizes the attention mechanism to defend against perturbations. Modelling node features as normal distributions, the R-GCN assigns attention scores according to the variances of the nodes.
\item {\bf Pro-GNN}\cite{prognn}: Exploring the low rank and sparsity properties of adjacency matrices, Pro-GNN combines structure learning with the GNN in an end-to-end manner.
\end{itemize}

\subsubsection{Baselines}
Since we propose a novel attack condition which is not explored in previous literature, we utilizes baselines from \cite{mettack} under our proposed constraints. All baselines are restricted by the $\phi(G)$ constraint.
\begin{itemize}
\item {\bf DICE-BB}: This baseline is the black-box version of the DICE (Disconnect Internally, Connect Externally) algorithm. No ground-truth labels are accessible and spectral clustering is utilized to generate pseudo-labels. 
\item {\bf Random}: We also use random attack as a baseline. This baseline randomly adds $\Delta$ edges in the graph.
\item {\bf Mettack}: We created a black-box version of Mettack. For fair comparison it also utilizes pseudolabels and the first partition of the nodes is used for training.
\end{itemize}

\begin{table*}
\centering
\begin{tabular}{c c | c c c c}
Dataset & Attack & 20\% & 40\% & 60\% & 80\% \\ \hline
\multirow{4}{*}{Cora}
& BBGA & \textbf{23.90$\pm$1.28}
& \textbf{28.98$\pm$1.69}
& \textbf{33.48$\pm$2.01}
& \textbf{37.12$\pm$2.09} \\
& DICE-BB & 21.42$\pm$ 1.16 (-10.37\%)
	& 25.69$\pm$1.19 (-10.71\%)
	& 29.43$\pm$1.31 (-12.10\%)
	& 33.14$\pm$1.52 (-10.72\%) \\
& Random & 21.34$\pm$1.74 (-10.71\%)
	& 25.77$\pm$1.19 (-11.08\%)
	& 28.43$\pm$1.71 (-15.08\%)
	& 32.73$\pm$1.76 (-11.83\%) \\
& Mettack & 22.58$\pm$1.19 (-5.23\%)
	& 26.82$\pm$1.57 (-7.45\%)
	& 28.88$\pm$1.26 (-13.74\%)
	& 32.71$\pm$1.51 (-11.88\%) \\ \hline
\multirow{4}{*}{Citeseer}
& BBGA & \textbf{29.44$\pm$1.22}
	& \textbf{32.17$\pm$1.15}
	& \textbf{35.69$\pm$1.24}
	& \textbf{39.46$\pm$2.01} \\
& DICE-BB & 27.92$\pm$0.85 (-5.16\%)
	& 30.85$\pm$1.41 (-4.10\%)
	& 33.91$\pm$1.53 (-4.99\%)
	& 37.22$\pm$2.21 (-5.68\%) \\
& Random & 27.25$\pm$1.10 (-7.44\%)
	& 29.96$\pm$1.48 (-6.87\%)
	& 32.12$\pm$0.85 (-10.00\%)
	& 33.68$\pm$1.52 (-14.65\%) \\
& Mettack & 27.19$\pm$0.76 (-7.64\%)
	& 30.55$\pm$1.29 (-5.04\%)
	& 32.59$\pm$1.69 (-8.69\%)
	& 36.70$\pm$1.67 (-6.99\%) \\ \hline
\multirow{4}{*}{Cora-ML}
& BBGA & \textbf{30.29$\pm$3.03}
& \textbf{45.10$\pm$3.58}
	& \textbf{52.69$\pm$5.26}
	& \textbf{60.36$\pm$4.96} \\
& DICE-BB & 25.81$\pm$2.47 (-14.79\%)
	& 34.10$\pm$4.35 (-24.39\%)
	& 42.27$\pm$7.37 (-19.78\%)
	& 55.77$\pm$5.60 (-7.60\%) \\
& Random & 24.79$\pm$1.64 (-18.16\%)
	& 30.78$\pm$2.40 (-31.75\%)
	& 40.74$\pm$3.79 (-22.68\%)
	& 47.74$\pm$3.90 (-20.91\%) \\
& Mettack & 28.07$\pm$1.79 (-7.33\%)
	& 36.81$\pm$3.11 (-18.38\%)
	& 42.96$\pm$2.93 (-18.47\%)
	& 46.92$\pm$2.93 (-22.27\%) \\ \hline
\end{tabular}
\caption{Misclassification rates when attacking against the GCN (Classification accuracy$\pm$Std).}
\label{selected}
\end{table*}

\begin{figure*}
\centering
\includegraphics[width=1.0\textwidth]{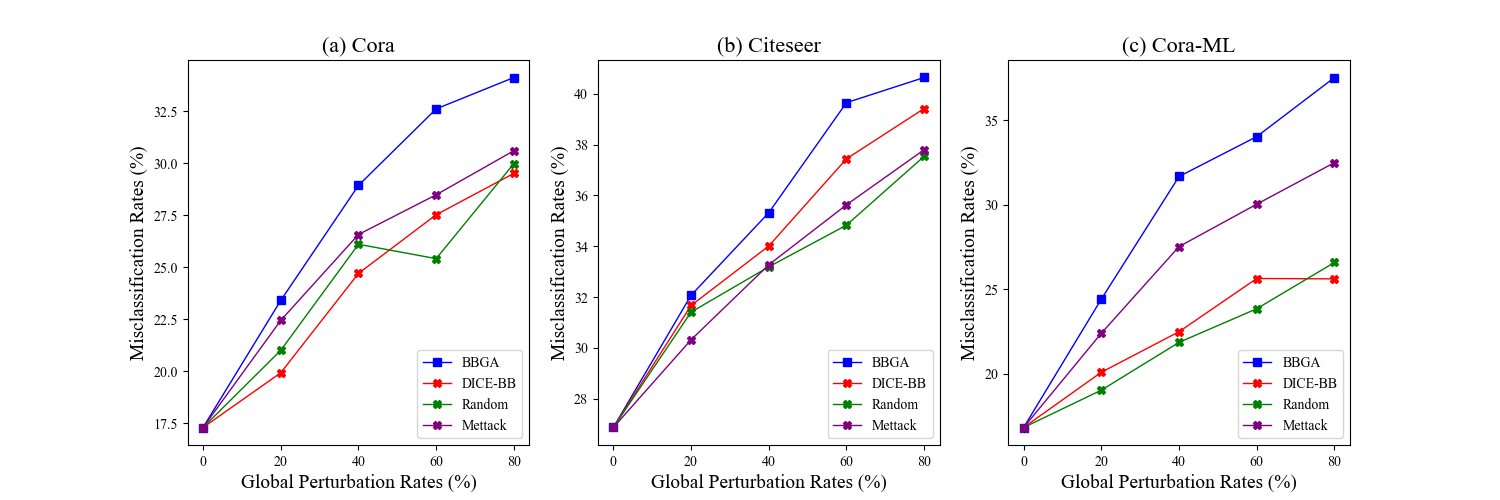}
\Description{Misclassification rates when attacking against GCN-Jaccard, (a) Cora, (b) Citeseer, (c) Cora-ML.}
\caption{Misclassification rates when attacking against GCN-Jaccard, (a) Cora, (b) Citeseer, (c) Cora-ML.}
\label{figjaccard}
\end{figure*}

\begin{figure*}
\centering
\includegraphics[width=1.0\textwidth]{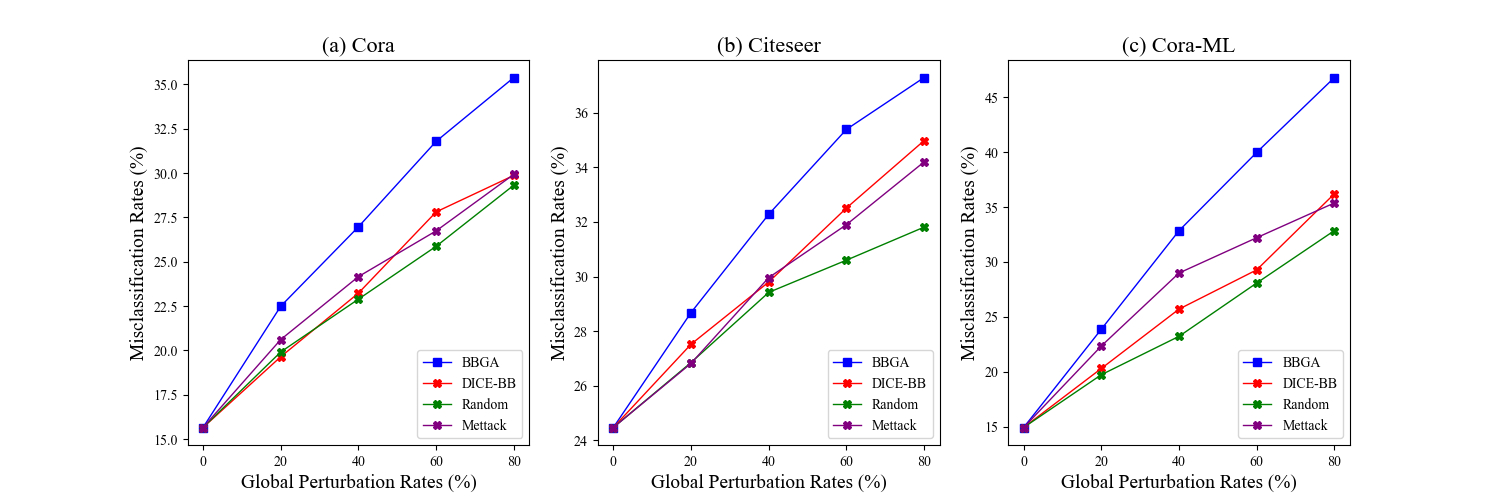}
\Description{Misclassification rates when attacking against R-GCN, (a) Cora, (b) Citeseer, (c) Cora-ML.}
\caption{Misclassification rates when attacking against R-GCN, (a) Cora, (b) Citeseer, (c) Cora-ML.}
\label{figrgcn}
\end{figure*}

\begin{figure*}
\centering
\includegraphics[width=1.0\textwidth]{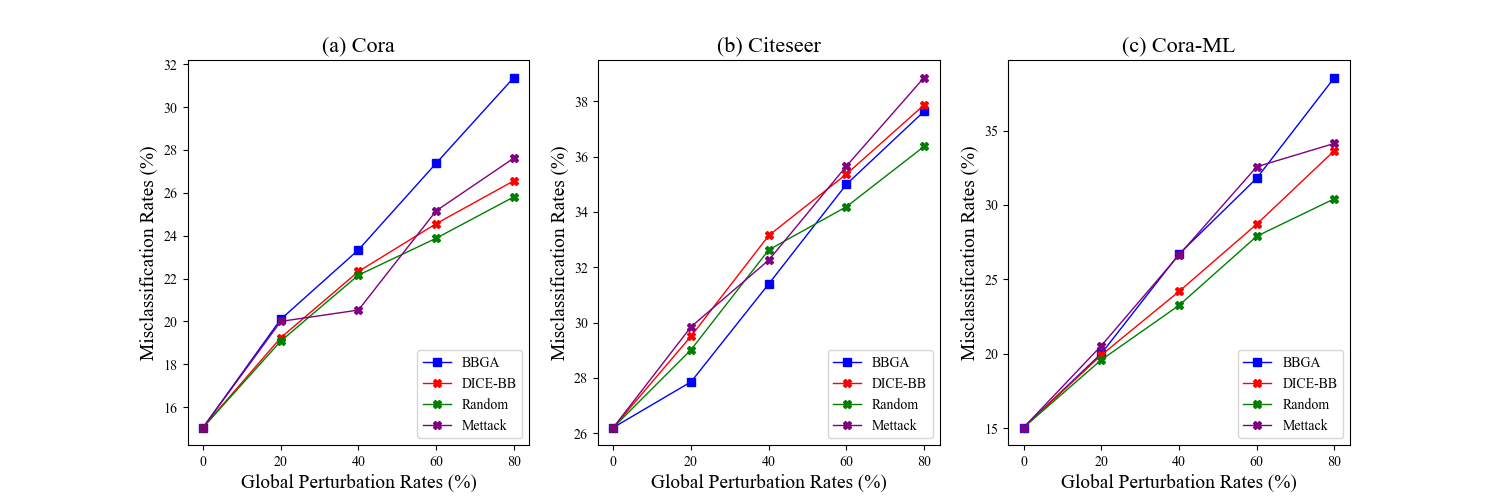}
\Description{Misclassification rates when attacking against Pro-GNN, (a) Cora, (b) Citeseer, (c) Cora-ML.}
\caption{Misclassification rates when attacking against Pro-GNN, (a) Cora, (b) Citeseer, (c) Cora-ML.}
\label{figprognn}
\end{figure*}

\subsubsection{Parameters and Experimental Settings}
We utilize scikit-learn \cite{scikit-learn} for spectral clustering with parameter $\gamma=0.001$. Following \cite{deeprobust} we set $\eta=0.01$. For other hyperparameters, we set $k=5$ and $T=100$.

We conducted our experiments on an Ubuntu 16.04 LTS server with two E5-2650 CPUs and 4 GTX 1080Ti GPUs. Python packages we used include Pytorch 1.5.0, Scikit-Learn 0.23.1, NumPy 1.18.5 and SciPy 1.3.1. Tensorflow 1.15.0 is used in R-GCN defenses.

For each perturbation rate we ran the experiments for 10 times. To verify that our proposed BBGA algorithm is not engaged to any specific training set, we randomly altered the dataset splits with scikit-learn\cite{scikit-learn} each time before the training of the defense algorithms. Following previous works \cite{mettack,prognn} we choose accuracy rate as the primary evaluation metric.

\subsection{Attack Performance}

In this subsection, we answer the research question {\bf RQ1}. Selected results when attacking against the original GCN is reported in Table \ref{selected}. 

We observe from the table that our proposed method outperforms the baselines in all situations when attacking against the original GCN model. Especially, BBGA outperforms the baselines by 20\% in several experimental settings.

The performance when attacking against various defend methods are reported in Figures \ref{figjaccard},  \ref{figrgcn} and \ref{figprognn}. As shown in the figures, our proposed BBGA algorithms achieves the best misclassification rates in most of the situations. The experiments show that our proposed method is able to achieve a higher misclassification rates when the training set is is not accessible.

\subsection{Case Study on Perturbations}

\begin{figure}
\centering
\includegraphics[width=0.4\textwidth]{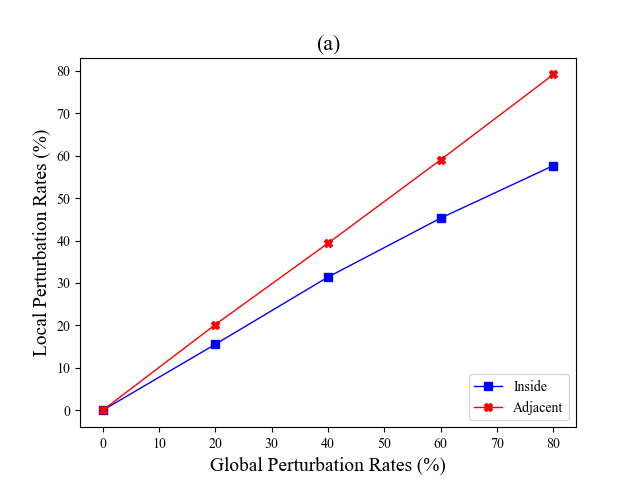}
\Description{The case study on the distribution of modifications by BBGA algorithm.}
\caption{The case study on the distribution of modifications by BBGA algorithm.}
\label{fig02}
\end{figure}

In this subsection, we answer the research question {\bf RQ2} via analysing the perturbations. For a random split of the Cora dataset, we reveal the local perturbation rates in Figure \ref{fig02}. As illustrated in the figure, the perturbations are not biased toward the training set. This demonstrate the evenness of the modifications of our proposed method. It explains the reason why our BBGA algorithm works without accessing the training set.

\subsection{Ablation Study and Parameter Analysis}

\begin{figure}
\centering
\includegraphics[width=0.4\textwidth]{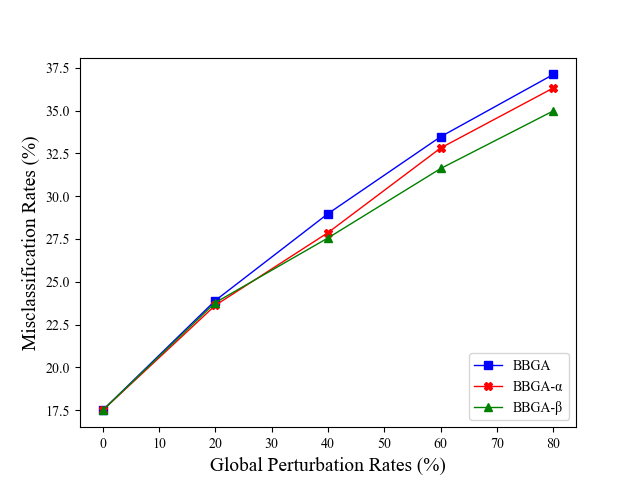}
\Description{Results of the ablation study}
\caption{Results of the ablation study.}
\label{fig03}
\end{figure}

To answer the research question {\bf RQ3}, we conducted ablation studies and parameter analysis. For ablation study, we created two variant of our method. BBGA-$\alpha$ removes the filter of low-variance node-pairs in Eq. \ref{score}. BBGA-$\beta$ removes the $k$-fold greedy choice procedure and in each step, exactly one partition is chosen randomly to compute the meta-gradient. We took Cora and the GCN model as an example. As revealed in Figure \ref{fig03}, BBGA has the best performance while BBGA-$\beta$ performs the worst. This indicates that both considering multiple partitions and filtering low-variance node-pairs help in increasing the attack performance.

\begin{figure}
\centering
\includegraphics[width=0.4\textwidth]{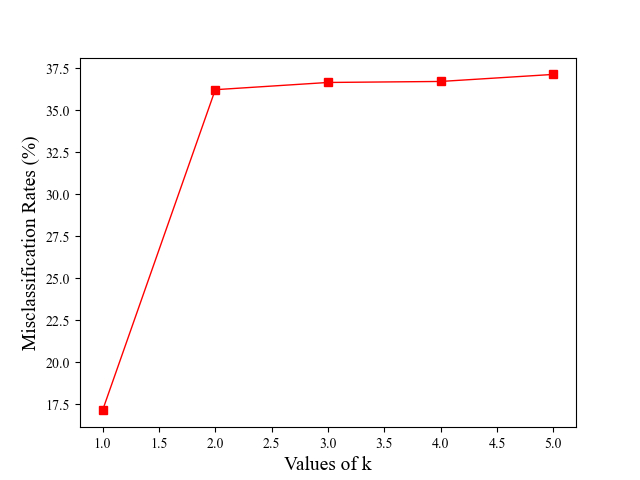}
\Description{Results of the parameter analysis on k.}
\caption{Results of the parameter analysis on $k$.}
\label{k_b}
\end{figure}

For parameter analysis, we varied the number of partitions $k$. Taking $80\%$ perturbation rates on Cora as an example, it is revealed that the attack performance grow as $k$ increases in general. However, since a larger $k$ implies a longer training time, we suggest to use the hyperparameter $k=5$. The results of the parameter analysis is reported in Figure \ref{k_b}.

\section{Conclusion}
Previous studies demonstrated that GCNs can be easily fooled by adversarial attacks. However, most existing adversarial attacks requires conditions that are not practical in real-world situations. In this paper, we demonstrate that utilizing training set in adversarial attacks, which is considered to be unrealistic, leads to the lack of robustness. By utilizing pseudo-labels and the $k$-fold greedy strategy, we propose the novel Black-Box Gradient Attack (BBGA) algorithm. Experimental results demonstrate that our proposed algorithm achieves promising performance as a black-box structure-oriented attack. To the best of our knowledge, this is the first gradient-based black-box graph attack and the first non-targeted structure attack that doesn't require any inquiry.  Future directions include further utilization of gradients in black-box scenarios and further investigations in black-box structural attacks.

\bibliographystyle{ACM-Reference-Format}
\bibliography{bbga}

\end{document}